%% file: main.tex
\documentclass[10pt,twocolumn,letterpaper]{article}

\usepackage{cvpr}              


\definecolor{cvprblue}{rgb}{0.21,0.49,0.74}
\usepackage[pagebackref,breaklinks,colorlinks,allcolors=cvprblue]{hyperref}

\usepackage{lmodern}
\usepackage{float}
\usepackage{array}

\def\paperID{18171} 
\def\confName{CVPR}
\def\confYear{2025}

\usepackage{bbding}
\usepackage{graphicx}
\usepackage{epstopdf}
\usepackage{epsfig}
\usepackage{booktabs}
\usepackage{pifont}

\newcommand\nonumfootnote[1]{%
\begingroup%
    \renewcommand\thefootnote{}\footnote{\hspace{-3.7pt}#1}%
    \addtocounter{footnote}{-1}%
\endgroup%
}

\begin{document}
\title{SCIGS: 3D Gaussians Splatting from a Snapshot Compressive Image}


\author{Zixu Wang$^{1,2,3,*}$\qquad Hao Yang$^{1,2,4,*}$\qquad Yu Guo$^{1,2,4,\dagger}$\qquad Fei Wang$^{1,2,4}$ \\
$^{1}$National Key Laboratory of Human-Machine Hybrid Augmented Intelligence\\
$^{2}$National Engineering Research Center for Visual Information and Applications\\
$^{3}$School of Software Engineering, Xi'an Jiaotong University\\
$^{4}$Institute of Artificial Intelligence and Robotics, Xi'an Jiaotong University\\
{\tt\small \{zixuwang, cvlab.yang\}@stu.xjtu.edu.cn\qquad yu.guo@xjtu.edu.cn\qquad wfx@mail.xjtu.edu.cn}}


\twocolumn[{
\renewcommand\twocolumn[1][]{#1}
\maketitle
\centering
    \captionsetup{type=figure}
    \vspace{-40pt}
    \includegraphics[width=1.0\textwidth]{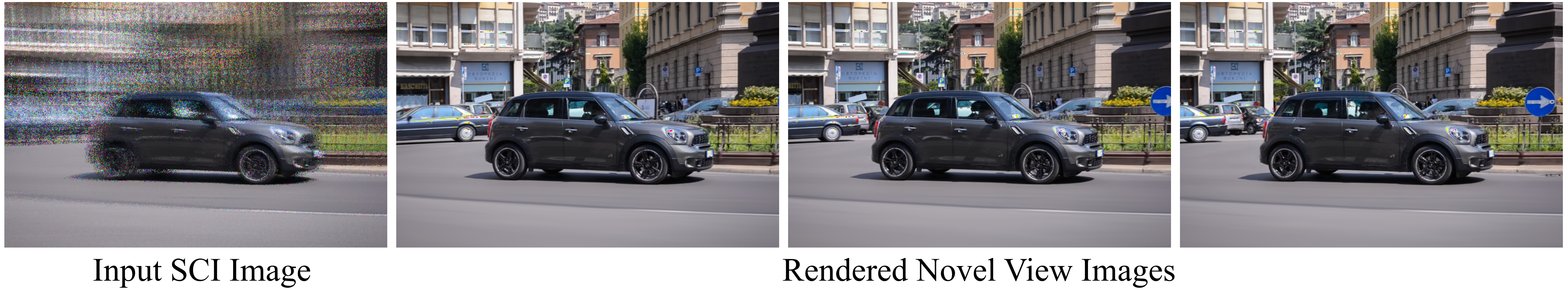}
    \captionof{figure}{Given a single compressed image of a dynamic scene as input, the proposed SCIGS can reconstruct a high-quality dynamic 3D scene and recover multi-view consistent images.}
    \vspace{10pt}
}]
\input{sec/0_abstract}

\input{sec/1_intro}

\input{sec/2_related_work}
\input{sec/3_method}
\input{sec/4_experiments}

\input{sec/5_conclusion}
{
    \small
    \bibliographystyle{ieeenat_fullname}
    \bibliography{main}
}

\input{sec/X_suppl}

\end{document}

%% file: sec/0_abstract.tex
\begin{abstract}
Snapshot Compressive Imaging (SCI) offers a possibility for capturing information in high-speed dynamic scenes, requiring efficient reconstruction method to recover scene information. 
Despite promising results, current deep learning-based and NeRF-based reconstruction methods face challenges: 1) deep learning-based reconstruction methods struggle to maintain 3D structural consistency within scenes, and 2) NeRF-based reconstruction methods still face limitations in handling dynamic scenes. 
To address these challenges, we propose SCIGS, a variant of 3DGS, and develop a primitive-level transformation network that utilizes camera pose stamps and Gaussian primitive coordinates as embedding vectors.
This approach resolves the necessity of camera pose in vanilla 3DGS and enhances multi-view 3D structural consistency in dynamic scenes by utilizing transformed primitives.
Additionally, a high-frequency filter is introduced to eliminate the artifacts generated during the transformation.
The proposed SCIGS is the first to reconstruct a 3D explicit scene from a single compressed image, extending its application to dynamic 3D scenes. 
Experiments on both static and dynamic scenes demonstrate that SCIGS not only enhances SCI decoding but also outperforms current state-of-the-art methods in reconstructing dynamic 3D scenes from a single compressed image.
The code will be made available upon publication.

\nonumfootnote{* Equal contribution \qquad $\dagger$ Corresponding author}
\end{abstract}
\vspace{-30pt}

%% file: sec/1_intro.tex
\section{Introduction}
\label{sec:intro}

High-speed imaging techniques are widely used in science research, sports, aerospace, etc. However, conventional high-speed imaging techniques typically entail significant expenditure on hardware and substantial storage requirements. Facing these challenges, Compressed Sensing (CS)\cite{candes2006robust, donoho2006compressed} and video Snapshot Compressive Imaging (SCI)\cite{yuan2021snapshot} technology has been developed. A SCI system usually has two components: a hardware encoder and a software decoder. 
During an exposure time, the hardware encoder uses multiple designed masks to divide an exposure process into multiple frames and modulate them into a compressed image. The software decoder can then use the masks to decode the compressed image into high frame rate images. 
This makes it possible for capturing high-speed video with ordinary cameras, which can reduce the hardware costs and storage costs.

For the hardware encoder of SCI systems, several mature approaches \cite{liu2018rank, llull2013coded} are proposed, though the decoding part still faces challenges. Existing decoding methods are categorized into model-based methods \cite{liao2014generalized, yang2014compressive, yuan2016generalized} and deep learning-based methods \cite{cheng2021memory, cheng2020birnat, meng2020gap, qiao2020deep, wang2023efficientsci, wang2022spatial}. Model-based methods use iterative optimization based on natural image priors, offering flexibility across resolutions and compression rates, but they suffer from the long processing time cost and low output quality. In contrast, deep learning-based methods leverage network architectures for end-to-end decoding of compressed images, achieving better real-time performance and higher image quality. However, both methods neglect the structure of the 3D scene, leading to inconsistencies across views. To address this, SCINeRF \cite{li2024scinerf} recovers 3D NeRF representations from a single compressed image by jointly optimizing NeRF and camera poses, yielding promising results. Yet, it performs suboptimally in dynamic scenes, common in high-speed photography. Additionally, NeRF-based reconstruction, with its large number of parameters in MLP, CNN, and Transformer architectures, demands significant training time and memory.

 In view of the fact that existing NeRF-based 3D reconstruction methods are incompetent for dynamic scene reconstruction and deep learning-based reconstruction methods struggle to maintain 3D structural consistency within scenes, inspired by the impressive representation capabilities and flexibility of 3DGS, to adapt dynamic scene, we propose SCIGS, the first method to construct a 3D explicit scene from a single compressed image and further extend this to dynamic 3D scenes. 
 Due to the inability to extract the initialization of the camera pose and Gaussians from the compressed image, and constrained by the discrete nature of 3D Gaussians, it is challenging to optimize the camera poses and 3D gaussians simultaneously. 
 to address this issue, a transformation network is proposed, which not only can decouple the transformation field from compressed image for adapting to the dynamic scenes, but also provide a solution for the oscillations of camera poses during optimization.
 A high-frequency filter is subsequently applied to suppress artifacts generated during transformation. 
 Extensive experiments both on static and dynamic scenes demonstrate that the proposed method achieves superior image quality in SCI decoding tasks and outperforms other methods for 3D scene reconstruction from compressed images, particularly in dynamic scenarios.

\noindent Our main contributions can be summarized as follows:
\begin{itemize}
    \item The proposed SCIGS is the first to recover explicit 3D representations from a single snapshot compressed image within the 3D Gaussian Splatting (3DGS) framework.

    \item Introducing camera pose stamps and a Gaussian primitive-level transformation network, we substitute the optimization of camera poses with a transformation of Gaussians, tackling the oscillations of camera poses during optimization equivalently.
    
    \item Extensive experiments demonstrate that SCIGS synthesizes high-quality novel-view images in both static and dynamic scenes, surpassing existing SCI image reconstruction methods.
\end{itemize}

%% file: sec/2_related_work.tex
\section{Related Work}
\label{sec:related_work}

In this paper, two main fields of the prior works are reviewed: the SCI image decoding methods and the 3D reconstruction methods.

{\bf SCI image decoding.}
Decoding a SCI image back to the original images is an ill-posed problem, to solve which, various types of priori knowledge are utilized in traditional methods, including total variation (TV)\cite{yuan2016generalized}, low-rank prior\cite{dong2014compressive, maggioni2012video, liu2018rank}, over-complete dictionary\cite{hitomi2011video, liu2013efficient}, Gaussian mixture model (GMM)\cite{yang2014video, yang2014compressive}, etc.
These traditional methods are flexible in different scenarios, but has high computational cost and poor reconstruction quality. 

With the development of deep learning, deep denoising networks are recognized as an effective image prior, based on which various plug-and-play (PnP) methods\cite{venkatakrishnan2013plug, yuan2020plug, qiao2021snapshot, wu2023adaptive} are proposed, which are able to achieve excellent performance while maintaining the flexibility of traditional methods. 
In addition, many end-to-end methods based on deep learning are also proposed, and these SCI decoders employ various network architectures\cite{cheng2021memory, wang2023efficientsci, cao2024hybrid} to extract the information in compressed image, including CNN\cite{qiao2020deep}, RNN\cite{cheng2020birnat}, Tranformer\cite{dosovitskiy2020image, wang2022spatial}, etc. 
To address the computational cost of the current mainstream methods based on Transformer, EfficientSCI\cite{wang2023efficientsci, cao2024hybrid} introduces hierarchical dense connections in residual block. 
Moreover, existing SCI image decoding methods do not consider the structure of the underlying 3D scene during reconstruction, resulting in the lack of multi-view consistency. 

\textbf{3D Reconstruction.}
Current mainstream works on 3D reconstruction mainly include NeRF-based reconstruction and 3DGS-based reconstruction. 

Mildenhall et al. proposed Neural Radiance Fields (NeRFs) \cite{mildenhall2021nerf}, which employs a Multi-Layer Perceptron (MLP) to implicitly learn 3D scene and renders images by Ray-Marching, has shown excellent performance compared to the previous methods, but its use of MLP leads to high train cost. 
Many NeRF-based variants have since emerged, such as Mip-NeRF360 \cite{barron2022mip}, which balances image quality and rendering speed. 
Other works focused on camera-free methods, like NeRF-\cite{wang2021nerf}, which jointly estimates scene and camera poses, and SCINeRF \cite{li2024scinerf}, which optimizes NeRF parameters and camera poses in static scenes. 
However, SCINeRF struggles with dynamic scene reconstruction due to NeRF’s limitations.

On the other hand, 3DGS \cite{kerbl20233d} uses 3D Gaussians for scene reconstruction, offering faster rendering and high image quality. 
Variants like 2DGS \cite{huang20242d} and Mip-Splatting \cite{yu2024mip} improve multi-view consistency and reduce rendering artifacts. 
Methods for handling scenarios without camera poses or in dynamic scenes have also been proposed, such as COLMAP-Free 3DGS \cite{fu2024colmapfree3dgaussiansplatting} and GS-SLAM \cite{yan2024gs}, which optimize camera poses and Gaussians iteratively. 
However, these methods struggle with large discrepancies between initial and target camera poses. 
iComMa \cite{sun2023icomma} addresses this with a matching loss for optimization guidance, and 6DGS \cite{bortolon20246dgs} estimates camera poses by reversing the 3DGS rendering process.
For dynamic scenes, methods like Deformable 3DGS \cite{yang2024deformable} introduce deformation fields, while 4D Gaussian Splatting \cite{wu20244d} extends this idea with neural voxel encoding to handle dynamic scene changes.

%% file: sec/3_method.tex
\begin{figure*}[htbp]
\centering
\includegraphics[width=1.0\textwidth]{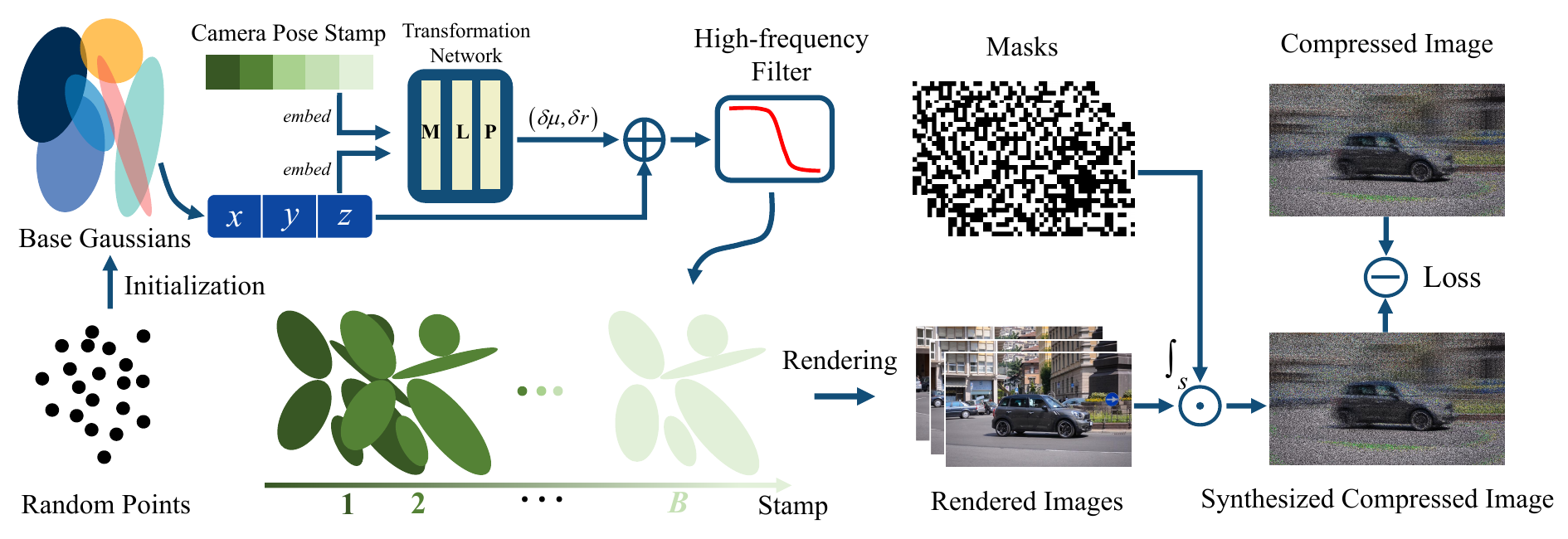}
\caption{\textbf{The pipeline of the proposed SCIGS.} Given a set of randomly initialized 3D Gaussians and a camera pose, and introducing the same number of camera pose stamps as the compression ratio, our transformation network takes the Gaussian primitives and the camera pose stamps as inputs, followed by a high-frequency filter, outputs 3D Gaussians under different camera pose stamps. These camera-pose-aware transformed 3D Gaussians are then rendered to images under the given camera viewpoint, and are modulated by a given set of masks to generate compressed images.}
\label{fig:pipeline}
\end{figure*}

\section{Method}
\label{sec:method}

The pipeline of the proposed method is shown in Fig.~\ref{fig:pipeline}.
The input to the proposed method is a single compressed image and a set of masks. 
From a random initial point cloud, a set of initial 3D Gaussians $\mathcal{G}\left( {\mu,r,s,\sigma } \right)$ are created, which are defined by position $\mu$, opacity $\sigma$, and a 3D covariance matrix $\Sigma$ derived from quaternions $r$ and scaling vectors $s$. Then a fixed viewpoint camera is defined by the random external parameters and the given internal parameters. 
The appearance of the Gaussians at each viewpoint is represented by spherical harmonics (SH).
In order to substitute camera pose transformation by the camera-pose-aware transformation of the 3D Gaussians and to adapt to the dynamic scene, a transformation network $\mathcal{F}$ is introduced, which takes the positions of each 3D Gaussians and a camera pose stamp as inputs, outputs transformation of Gaussians.
To eliminate the high-frequency artifacts generated during the transformation, a high-frequency filter follows, before the differential Gaussian rasterization pipeline that outputs intermediate frame images. 
Afterwards, simulating the modulation process of SCI system, the intermediate frame images are modulated into compressed images. 
Along with the adaptive density control of the Gaussians, the 3D Gaussians and the transformation network are simultaneously optimized by fast back-propagation. 

\subsection{3D Gaussian Splatting}

In this paper, the efficient differentiable rasterization pipeline proposed in \cite{kerbl20233d} is employed to render images from 3D Gaussians. As the rendering primitive for 3DGS, a 3D Gaussian is defined as:
\begin{equation}
    {\mathcal G}\left( \boldsymbol{x} \right) = \exp \left( { - \frac{1}{2}{{\left( {\boldsymbol{x} - \boldsymbol{\mu} } \right)}^\top}{\Sigma ^{ - 1}}\left( {\boldsymbol{x} - \boldsymbol{\mu} } \right)} \right),
    \label{eq:3dgs1}    
\end{equation}
where $\Sigma$ is parameterized as a combination of quaternion $r$ and a 3D scaling vector $s$, defined as:
\begin{equation}
    \Sigma = \boldsymbol{R}\boldsymbol{S}{\boldsymbol{S}^\top}{\boldsymbol{R}^\top}.
     \label{eq:3dgs2}
\end{equation}

In order to render 3D Gaussians to 2D images, it is necessary to project the 3D Gaussians onto a 2D imaging plane. The covariance of the resulting 2D Gaussians after projection can be approximated as:
\begin{equation}
    \Sigma'=\boldsymbol{J}\boldsymbol{W}\Sigma{\boldsymbol{W} ^\top}{\boldsymbol{J}^\top},
    \label{eq:3dgs3}
\end{equation}
where $\boldsymbol{J}$ denotes the Jacobian matrix for the affine approximation of the projection transformation, $\boldsymbol{W}$ stands for the view matrix transitioning from world coordinates to camera coordinates.

Subsequently, during the rendering process, the color of given pixel $p$ can be calculated through the alpha blending of $N$ ordered 2D Gaussians:
\begin{equation}
    \begin{split}
        C(p) &= \sum_{i\in{N}}T_{i}\alpha_{i}c_{i},\\
        \alpha_{i} &= \sigma_{i}e^{-\frac{1}{2}(p-\mu_{i})^{\top}\Sigma'(p-\mu_{i})},\\
        T_{i} &= \prod _{j=1}^{i-1}{(1-\alpha_{j})},
    \end{split}
    \label {eq:alpha_blending}
\end{equation}
where $c_{i}$ represents the Gaussian color derived from the spherical harmonic coefficients. 

\subsection{Snapshot Compressive Imaging Model}

In the process of capturing compressed images in the SCI system, an exposure time is divided into $B$ time intervals by the corresponding $B$ encoding masks. 
Within each time interval, each pixel's exposure is determined by the value at the corresponding position on the respective mask, and the image sensor accumulates the exposure data from each pixel in that time interval onto the compressed image, resulting in a final compressed image.
Additionally, in the process of randomly generating binary masks, The probability of selecting a position on the mask for exposure is fixed, denoted as the Overlap Ratio (OR), which is selected through ablation experiments.
The entire imaging process can be formulated as follow:
\begin{equation}
    Y = \sum_{i=1}^{B}X_{i} \odot M_{i} + Z_{i},
    \label{eq:sci1}
\end{equation}
where $Y$, $X_{i} \in \mathbb{R}^{H\times W}$ are the modulated compressed image and the $i^{th}$ virtual image within exposure time respectively, $B$ denotes the temporal Compression Ratio(CR), $\odot$ denotes element-wise multiplication, and $Z\in \mathbb{R} ^ {H \times W}$ is the measurement noise. 
Additionally, this process is entirely differentiable.

\subsection{Transformation Network}

The proposed method takes a compressed image as input, making it impractical to extract camera poses and point clouds by COLMAP\cite{schonberger2016structure}, which necessitates optimizing camera poses and 3D Gaussians jointly.

When directly optimizing the camera by gradient descent, The gradient of the loss $\mathcal{L}_{c}$ on the camera extrinsic matrix $E_{c}$ is shown as follow:
\begin{equation}
    \begin{split}
        \frac{\partial{\mathcal{L}_{c}}}{\partial E_{c}} &= 
        \sum_{i=1}^{H \times W}\frac{\partial{\mathcal{L}_{c}}}{\partial{C}} 
        \frac{\partial{C}}{\partial{E_{c}}} 
        \\&= \sum_{i=1}^{H \times W}\frac{\partial{\mathcal{L}_{c}}}{\partial C}
        \sum_{j=1}^{M}\left(
            \frac{\partial{C}}{\partial{c_{j}}}\frac{\partial{c_{j}}}{\partial{E_{c}}} +
            \frac{\partial{C}}{\partial{\alpha_{j}}}\frac{\partial{\alpha_{j}}}{\partial{E_{c}}}
        \right),
    \end{split}
    \label{eq:TN_gradE}
\end{equation}
where $C$ and $c_{j}$ denote the color of the rendered pixel and the color of the Gaussian $\mathcal{G}_{j}$ respectively. $M$ indicates the number of visible 3D Gaussians.
\begin{figure*}[htbp]
\centering
\includegraphics[width=2.0\columnwidth]{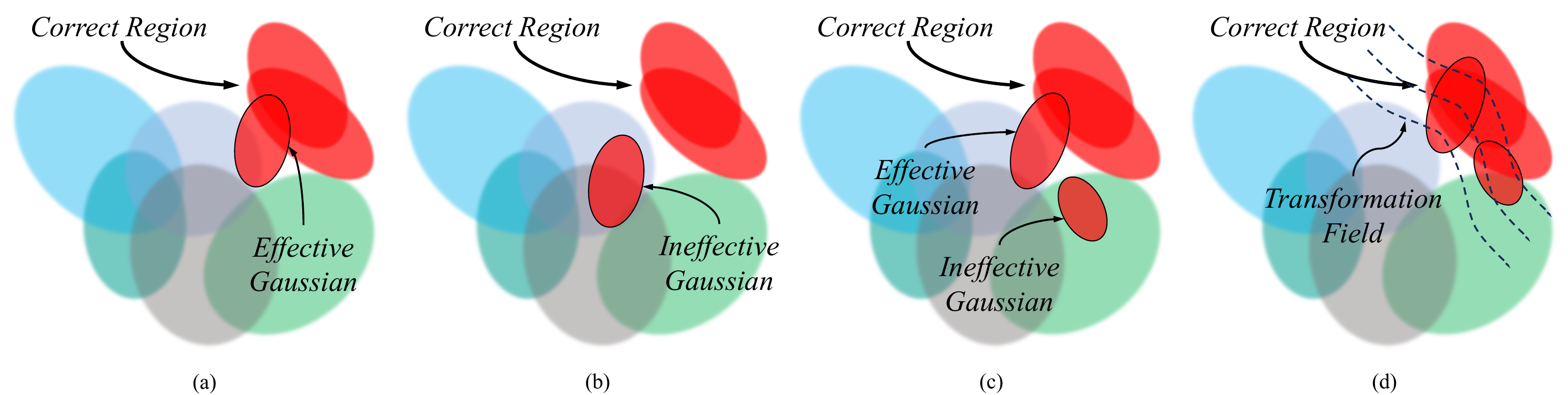}
\caption{(a) illustrates an effective Gaussian, (b) illustrates an ineffective Gaussian. (c) and (d) show how the transformation network converts ineffective Gaussians to effective Gaussians.}
\label{fig:overlap}
\end{figure*}

\begin{figure}[htbp]
\centering
\includegraphics[width=\columnwidth]{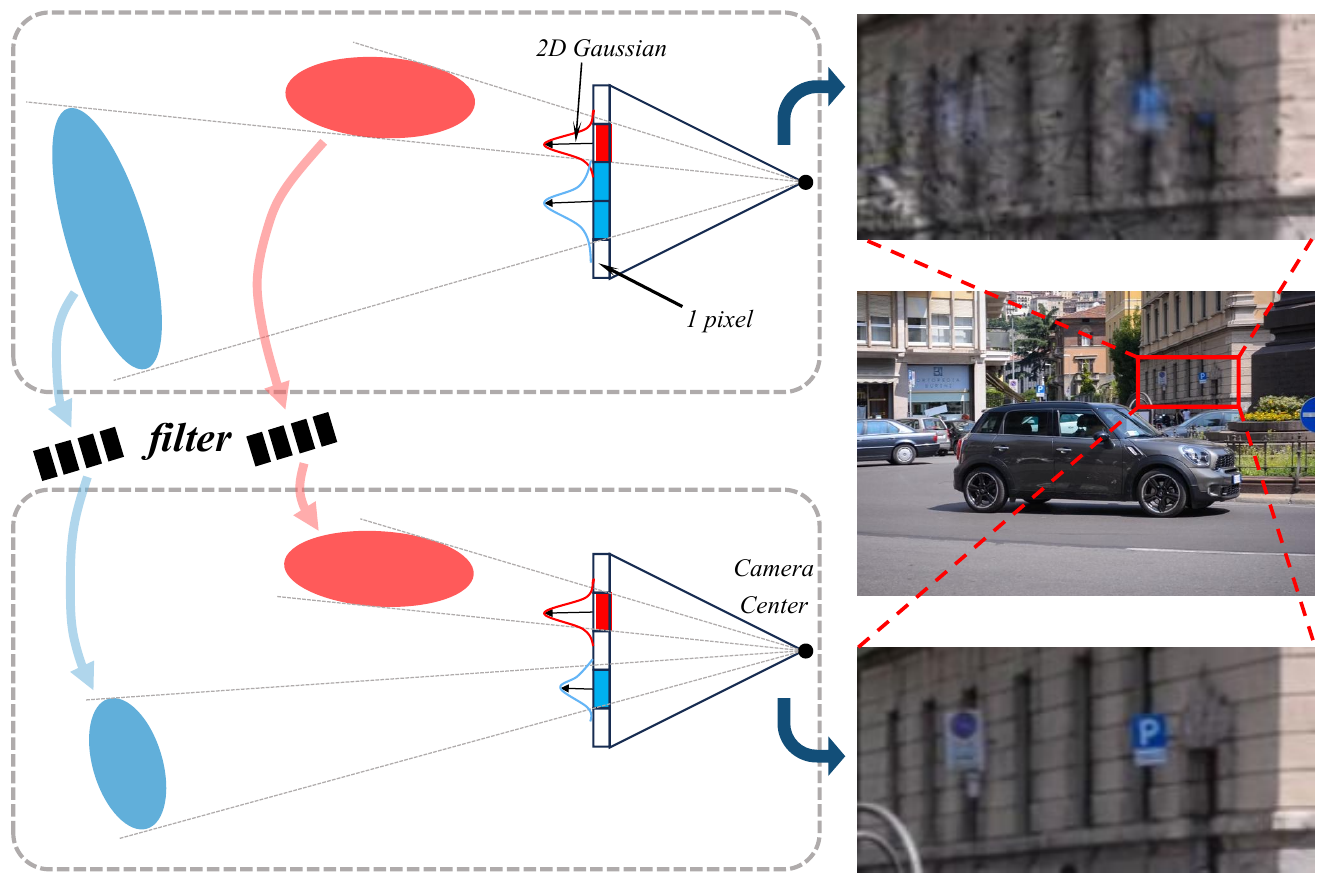}
\caption{\textbf{The illustration of the principle of the proposed high-frequency filter.} The Gaussians that cause high-frequency artifacts are filtered to eliminate the artifacts.}
\label{fig:filter}
\end{figure}

As shown in Fig.~\ref{fig:overlap}(a) and Fig.~\ref{fig:overlap}(b), the possible conditions of 3D Gaussians can be roughly divided into two types: the projection of the Gaussian overlapping with the correct region and the projection of the Gaussian not overlapping with the correct region, hereinafter referred to as the effective Gaussians and the ineffective Gaussians respectively.
 
When the Gaussian $\mathcal{G}_{i}$ is an ineffective Gaussian, it is worth noting that the gradients $\frac{\partial C}{\partial c_{j}}$ and $\frac{\partial C}{\partial \alpha_{j}}$  provide no guidance, which lead the camera optimized in chaotic direction according to Eq.~\ref{eq:TN_gradE}. 
In cases where there is a significant difference between the target camera pose and the initial camera pose, the number of ineffective Gaussian far exceeds the effective ones, hindering the oscillations of camera pose.

To avoid the above issue in camera pose optimization, this paper approaches the problem from the perspective of transforming Gaussian primitives and introduces a camera-pose-aware transformation network, with a multi-layer perceptron (MLP) as its core component. 
As formulated in Eq.~\ref{eq:TN}, The input of this network consists of the positions of Gaussians and a camera pose stamp, while the output is the increments of positions and quaternions, i.e., $\delta \mu$, $\delta r$ .
Since moving the initial camera to the correct pose is equivalent to moving the 3D Gaussians to the correct position in front of the camera.
Following the prior of local smoothness in natural images, contiguous Gaussians often have similar colors. With the presence of both effective and ineffective Gaussians in a neighborhood, as shown in Fig.~\ref{fig:overlap}(c), the correct gradients from effective Gaussians guide the transformation network optimized towards correct direction. Due to the continuity of the MLP, as shown in Fig.~\ref{fig:overlap}(d), nearby ineffective Gaussian points are also moved towards the correct direction with the transformation field, and gradually transformed into effective Gaussians. This process ultimately leads the 3D Gaussians to move towards the correct pose.
\begin{equation}
    \left( \delta \mu, \delta r \right) = \mathcal{F}\left( embed(\mu), embead(stamp)\right),
    \label{eq:TN}
\end{equation}
\begin{equation}
    embed(\mu) = \left( sin (2^{k}\pi\mu), cos (2^{k}\pi\mu)\right)^{L-1}_{k=0}.
    \label{eq:TN_embed}
\end{equation}


Benefiting from the transformation network acting directly on Gaussian primitives instead of camera poses, the Gaussian primitives can undergo different transformations under varying camera poses. This approach offers the transformation network a potential means to learn the movement of objects within the scene, enabling SCIGS to reconstruct dynamic scenes from a single SCI image.

\begin{table*}[t]
\begin{center}
\huge
\renewcommand{\arraystretch}{1.2} 
\resizebox{\textwidth}{!}
{
    \begin{tabular}{c|ccc|ccc|ccc|ccc|ccc|ccc}
        \toprule
            & \multicolumn{3}{c|}{Cozy2room} & \multicolumn{3}{c|}{Tanabata} & \multicolumn{3}{c|}{Factory} & \multicolumn{3}{c|}{Vender} & \multicolumn{3}{c|}{Airplants} & \multicolumn{3}{c}{Hotdog} \\
            & PSNR$\uparrow$ & SSIM$\uparrow$ & LPIPS$\downarrow$ & PSNR$\uparrow$ & SSIM$\uparrow$ & LPIPS$\downarrow$ & PSNR$\uparrow$ & SSIM$\uparrow$ & LPIPS$\downarrow$ & PSNR$\uparrow$ & SSIM$\uparrow$ & LPIPS$\downarrow$ & PSNR$\uparrow$ & SSIM$\uparrow$ & LPIPS$\downarrow$ & PSNR$\uparrow$ & SSIM$\uparrow$ & LPIPS$\downarrow$ \\
        \hline
        GAP-TV\cite{yuan2016generalized} & 21.77 & .4321 & .6031 & 20.42 & .4264 & .6250 & 24.05 & .5666 & .5149 & 20.00 & .3678 & .6882 & 22.85 & .4057 & .4986 & 22.35 & .7663 & .3179 \\
        PnP-FFDNet\cite{yuan2020plug} & 28.98 & .8916 & .0984 & 29.17 & .9032 & .1197 & 31.75 & .8977 & .1142 & 28.70 & .9235 & .1315 & 27.79 & .9117 & .1817 & 29.00 & \underline{.9765} & .0511 \\
        PnP-FastDVDNet\cite{yuan2021plug} & 30.19 & .9132 & .0793 & 29.73 & .9333 & .0980 & 32.53 & .9165 & .1055 & 29.68 & .9395 & .1043 & 28.18 & .9092 & .1757 & 29.93 & .9728 & .0522 \\
        EfficientSCI\cite{wang2023efficientsci} & 31.47 & \underline{.9327} & .0476 & 32.30 & \underline{.9587} & .0600 & 32.87 & .9259 & .0709 & 33.17 & .9401 & .0456 & \underline{30.13} & \textbf{.9425} & \underline{.1129} & \underline{30.75} & .9568 & \underline{.0461} \\
        SCINerf\cite{li2024scinerf} & \underline{33.23} & \textbf{.9492} & \underline{.0445} & \underline{33.61} & \textbf{.9638} & \underline{.0374} & \underline{36.60} & \underline{.9638} & \textbf{.0221} & \textbf{36.40} & \textbf{.9840} & \underline{.0298} & \textbf{30.69} & \underline{.9335} & \textbf{.0728} & \textbf{31.35} & \textbf{.9878} & \textbf{.0310} \\
        \hline
        SCIGS(ours) & \textbf{33.78} & .9191 & \textbf{.0423} & \textbf{35.12} & .9580 & \textbf{.0271} & \textbf{37.75} & \textbf{.9646} & \underline{.0291} & \underline{36.00} & \underline{.9641} & \textbf{.0192} & 27.18 & .7267 & .3003 & 29.31 & .9369 & .0809 \\
        \bottomrule
    \end{tabular}
}
\caption{\textbf{Quantitative SCI image reconstruction comparisons on the static datasets.} The results demonstrate that our method outperforms or approaches the existing image reconstruction methods and 3D reconstruction methods for SCI image on most datasets from static scenes. The best results are shown in bold and the second-best results are underlined.}
\label{tab:static}
\end{center}
\end{table*}

\begin{table*}[t]
\centering
\renewcommand{\arraystretch}{1.2} 
\resizebox{\textwidth}{!}{
    \begin{tabular}{c|ccc|ccc|ccc|ccc|ccc}
        \toprule
            & \multicolumn{3}{c|}{Bear} & \multicolumn{3}{c|}{Roundabout} & \multicolumn{3}{c|}{Turn} & \multicolumn{3}{c|}{Flamingo} & \multicolumn{3}{c}{Dance} \\
            & PSNR$\uparrow$ & SSIM$\uparrow$ & LPIPS$\downarrow$ & PSNR$\uparrow$ & SSIM$\uparrow$ & LPIPS$\downarrow$ & PSNR$\uparrow$ & SSIM$\uparrow$ & LPIPS$\downarrow$ & PSNR$\uparrow$ & SSIM$\uparrow$ & LPIPS$\downarrow$ & PSNR$\uparrow$ & SSIM$\uparrow$ & LPIPS$\downarrow$ \\
        \hline
        EfficientSCI\cite{wang2023efficientsci} & 26.81 & .8759 & .1040 & 22.08 & .7854 & .2934 & 22.30 & .7763 & .3613 & 25.97 & .8795 & .1386 & 24.83 & .9055 & .6268 \\
        SCINerf\cite{li2024scinerf} & 26.57 & .7974 & .1192 & 26.02 & .8394 & .1265 & 25.68 & .6596 & .2330 & 26.78 & .7954 & .1207 & 22.78 & .6960 & .2737 \\
        \hline
        SCIGS(ours) & \textbf{30.44} & \textbf{.9137} & \textbf{.0548} & \textbf{31.07} & \textbf{.9222} & \textbf{.0729} & \textbf{31.78} & \textbf{.8951} & \textbf{.0953} & \textbf{31.33} & \textbf{.9022} & \textbf{.0533} & \textbf{27.89} & \textbf{.9096} & \textbf{.0580} \\
        \bottomrule
    \end{tabular}
}
\caption{\textbf{Quantitative SCI image reconstruction comparisons on the dynamic datasets.} The results demonstrate that our method surpasses the current image reconstruction methods and 3D reconstruction methods for SCI image on all of datasets from dynamic scenes. The best results are shown in bold.}
\label{tab:dynamic}
\end{table*}

\subsection{High-frequency Filter}

When projecting 3D gaussians on imaging plane, a fixed 2D dilation factor is employed to ensure the projected 2D Gaussians are larger than one pixel, which leads to a systematically underestimation of scale.
Then, in the case of moving 2D Gaussians closer, the rendered Gaussian is thinner than they actually appear, which exhibits high-frequency artifacts on rendered image.
As shown in Fig.~\ref{fig:filter}, in the proposed frame, the transformation network moves the base Gaussians to the appropriate position, and similar to the above phenomenon, when optimizing based on the positions under a certain stamp, Gaussians under another stamp may be rendered as high-frequency artifacts. 
To address this issue, inspired by Mip-Splatting\cite{yu2024mip}, a high-frequency filter is introduced to eliminate the high-frequency artifacts.

Consider the rendering process as a sampling of 3D Gaussians, according to the Nyquist-Shannon sampling theorem: to recover the original signal from the sampled signal without distortion, the sampling frequency should be greater than two times the highest frequency of the signal.
For a camera with focal $f$, its sampling interval is one pixel in screen space, the sampling interval $\hat{T}$ and sampling frequency $\hat{v}$ for an object in depth $d$ in the camera coordinate are shown in Eq.~\ref{eq:HF_sample}.
\begin{equation}
    \hat{T} = \frac{1}{\hat{v}} = \frac{d}{f}.
    \label{eq:HF_sample}
\end{equation}

According to the sampling theorem, a primitive smaller than $2\hat{T}$ may result in artifacts during the splatting process, since the sampling frequency is below twice the signal frequency. 
Then the maximal sampling frequency of Gaussian $\mathcal{G}_{i}$ can be calculated by 
\begin{equation}
    \hat{v}_{i} = max\left(\left\{ \frac{f_{s}}{d_{s}}\right\}_{s=1}^{S}\right),
\end{equation}
where $S$ denotes the total number of camera pose stamps.
Given the maximal sampling frequency, as shown in Eq.~\ref{eq:HF_filter}, the high-frequency filter is achieved by convolving a low-pass filter Gaussian with 3D Gaussian before the rasterization pipeline.
\begin{equation}
    \begin{split}
        &\mathcal{G}_{i}' (x) = \left( \mathcal{G}_{i} \ast \mathcal{G}_{low}\right)(x) \\
        &= \sqrt{\frac{\lvert \Sigma_{i}\rvert}{\lvert \Sigma_{i} + \frac{\gamma}{\hat{v}_{i}} \cdot \textbf{I}\rvert}} 
        e^{\left( { - \frac{1}{2}{{\left( {\boldsymbol{x} - \boldsymbol{\mu_{i}} } \right)}^\top}{( \Sigma_{i} + \frac{\gamma}{\hat{v}_{i}} \cdot \textbf{I} ) ^{ - 1}}\left( {\boldsymbol{x} - \boldsymbol{\mu_{i}} } \right)} \right)},
    \end{split}
    \label{eq:HF_filter}
\end{equation}
where $\gamma$ is a hyperparameter controlling the filter size. the scale of the low-pass filtered Gaussian is $\frac{\gamma}{\hat{v}_{i}}$, which ensures that the scale of the filtered 3D Gaussian is not larger than the sampling interval after convolution.
In the additional experiments, an ablation experiment demonstrated the validity of the high-frequency filter.

%% file: sec/4_experiments.tex
\begin{figure*}[t]
\centering
\includegraphics[width=1.0\textwidth]{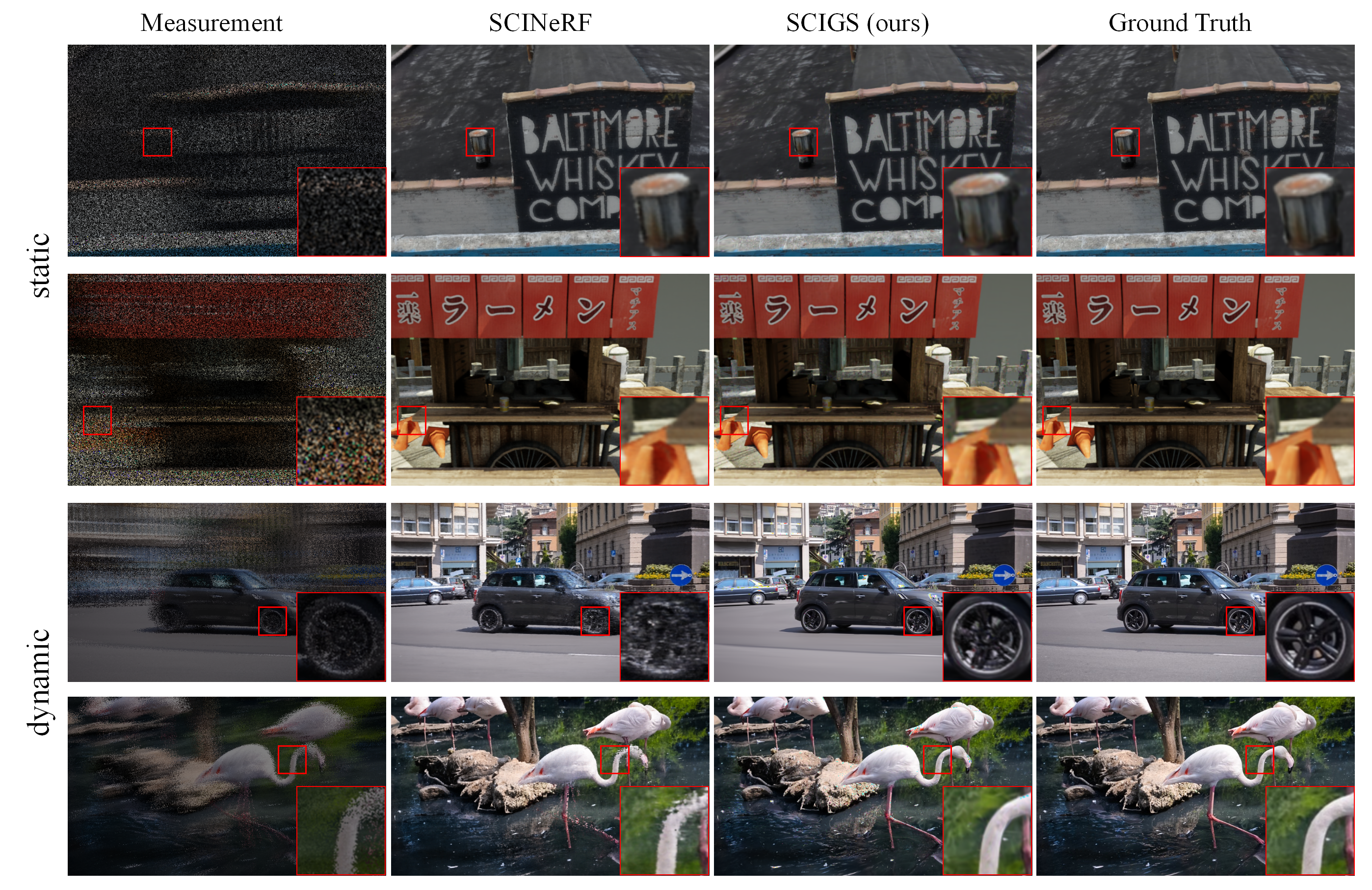}
\caption{\textbf{Qualitative evaluations on the synthetic dataset compare our proposed method (SCIGS) with the SOTA SCI image method (SCINeRF).}  From top to bottom are two static scenes (factory and tanabata) and two dynamic scenes (roundabout and flamingo). The experiments show that our method achieves comparable image recovery performance from a single compressed image in static scenes, while demonstrating superior performance in dynamic scenes.}
\label{fig:compare}
\end{figure*}

\section{Experiments}
\label{sec:experiment}

To evaluate the effectiveness of this method, extensive experiments were conducted using compressed images from both static and dynamic scenes. 
The results demonstrate that the proposed method delivers higher performance compared with existing works in SCI image decoding task and enhance the image quality in 3D reconstruction from compressed image.

\subsection{Experiment Setup}

\textbf{Datasets.}
For fair evaluation of the proposed SCIGS, we follow SCINeRF and use six static scenes, including $Airplants$ in LLFF\cite{mildenhall2019local} with solution $512 \times 512$, $Hotdog$ in NeRF Synthetic360\cite{mildenhall2021nerf} with solution $400 \times 400$ and four datasets generated from the scenes in DeblurNeRF\cite{ma2022deblur} ($Cozy2room$, $Tanabata$, $Factory$, and $Vender$) with solution $400 \times 600$. For dynamic scenes, we use five dataset from DAVIS2017\cite{pont20172017} with resolution $480 \times 894$(480p). 

\noindent \textbf{Baseline methods and evaluation metrics.}
In the comparison experiments, the SOTA SCI image decoding methods and SOTA reconstruction method for compressed image are used for comparison, including GAP-TV\cite{yuan2016generalized}, PnPFFDNet\cite{yuan2020plug}, PnP-FastDVDNet\cite{yuan2021plug}, EfficientSCI\cite{wang2023efficientsci}, and SCINeRF\cite{li2024scinerf}. Widely used metrics are employed for quantitative evaluations, including the structural similarity index(SSIM), peak signal-to-noise-ratio(PSNR), and learned perceptual image patch similarity(LPIPS)\cite{zhang2018unreasonable}.

\noindent \textbf{Implementation details.}
We use PyTorch framework with NVIDIA RTX A6000 GPUs for training. We used two independent Adam optimizers for the 3D Gaussians and the transformation network , and set the number of layers $L = 6$ in position coding function of the transformation network, the depth of the MLP $D = 8$ and the dimension of the hidden layer $W = 512$.

\subsection{Result and Analysis}

\textbf{Static scenes.}
The existing SOTA SCI image decoding methods and SCINeRF are compared with the proposed method on static datasets. As shown in Table~\ref{tab:static} and Fig.~\ref{fig:compare}, in the majority of datasets, our method surpasses the SOTA SCI image decoding methods and exceeds or approaches the image quality of SCINeRF. This empirical result demonstrates the efficacy of our SCIGS in recovering 3D scenes from compressed images. It is noteworthy that our approach did not outperform SCINeRF and EfficientSCI in $airplants$ and $hotdog$, which can be attributed to the fact that our method indirectly decodes the compressed image by recovering 3D representation, and the detail information lost due to the lack of texture information in $airplants$ and the large-scale camera movements in $hotdog$.

\noindent \textbf{Dynamic scenes.}
We also compared our method against SOTA SCI
image restoration methods (EfficientSCI) and the SOTA 3D reconstruction method from compressed images (SCINeRF) in dynamic scenes. As shown in Table~\ref{tab:dynamic} and Fig.~\ref{fig:compare}, in all scenes, our SCIGS outperforms SCINeRF on all metrics. This result demonstrates the efficacy of SCIGS in recovering dynamic scenes from compressed images. Qualitatively, as shown in Fig.~\ref{fig:compare}, we note that SCIGS demonstrates excellent performance in recovering the details of the moving objects in the scene, where existing methods fails to restore.

\begin{figure*}[t]
\centering
\includegraphics[width=1.0\textwidth]{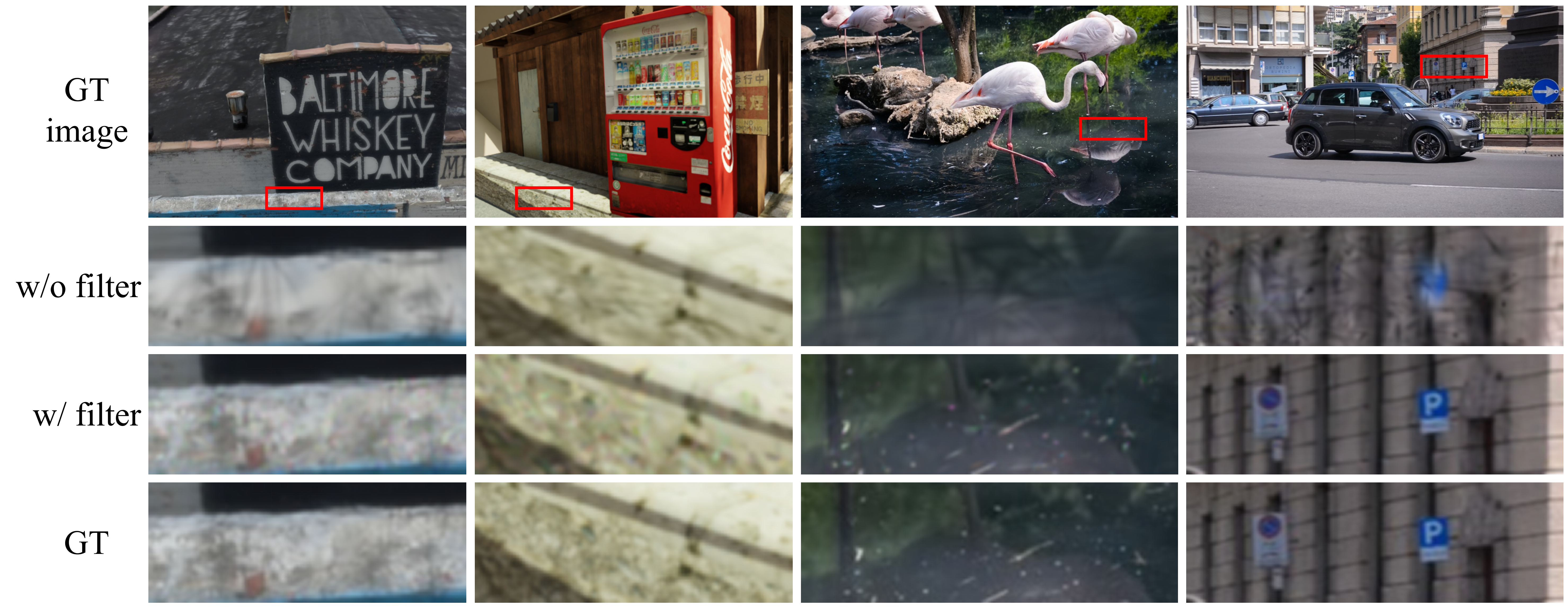}
\caption{\textbf{Ablation study on the high-frequency filter of SCIGS in both static and dynamic scenes.} The first two columns represent static scenes (factory and vender), while the last two columns show dynamic scenes (flamingo and roundabout). As shown, the addition of the high-frequency filter significantly enhances image quality in both static and dynamic scenes, with reduced artifacts and clearer details, further validating the effectiveness of the high-frequency filter across various scene types. }
\label{fig:ablation}
\end{figure*}

\subsection{Additional Study}

{\bf Mask overlapping rate. } 
We assess the impact of various mask overlapping rates during the SCI image modulated. The mask overlapping rate is defined by the probability that a mask selects a specific pixel for exposure, which is formulated as follow:
\begin{equation}
    OR(x,y) = \frac{\sum^{N}_{i=1}M_{i}(x,y)}{N}
    \label{eq:or}
\end{equation}
where $OR$ denotes mask overlapping rate, $M_{i}$ indicates $i$-th mask and $N$ is the number of Intermediate frame.
From Eq.~\ref{eq:or}, lower the mask crossing rate means the sparser sampling, which result in the less image information retained, and conversely, the denser the sampling leads to the more information retained. However, too high a sampling rate will increase the ambiguity of the compressed image and may result in a blurred decoded image. As shown in Table~\ref{tab:or}, we tested different overlapping rates on multiple datasets, and the results showed that the image quality first increased and then decreased when the overlapping rate increased from 0.125 to 0.25, and decreased after 0.25. Empirically, we selected overlap rate of all experiments within 0.25.

\begin{table}[t]
    \centering
    \footnotesize 
    \renewcommand{\arraystretch}{1.3} 
    \begin{tabular}{>{\centering\arraybackslash}p{0.13\columnwidth} | >{\centering\arraybackslash}p{0.22\columnwidth}  >{\centering\arraybackslash}p{0.22\columnwidth}  >{\centering\arraybackslash}p{0.22\columnwidth}}
        \toprule
        OR & PSNR$\uparrow$ & SSIM$\uparrow$ & LPIPS$\downarrow$ \\
        \hline
         0.125 & 30.32 & .9066 & .0634 \\
         0.25 & 30.41 & .8954 & .0814 \\
         0.5 & 29.04 & .8569 & .1145 \\
         0.75 & 27.50 & .8294 & .1336 \\
        \bottomrule
    \end{tabular}
    \caption{\textbf{The average metrics of image quility in the additional study on mask overlapping rate.} the quality of reconstruction increases first and then decreases with the overlapping rate ranging from 0.125 to 0.75.}
    \label{tab:or}
\end{table}

\subsection{Ablation Experiment}

To validate the efficacy of the proposed high-frequency filter, we conducted ablation experiments to compare the performance of our method before and after incorporating the filter, evaluated across both static and dynamic scenes. The results, presented in Table~\ref{tab:ablation}, demonstrate that the inclusion of the high-frequency filter significantly improves the quality of the recovered images, particularly in terms of visual fidelity and artifact reduction.

Moreover, the effectiveness of the high-frequency filter is evident in the qualitative experiments. As shown in Fig.~\ref{fig:ablation}, the filter almost completely eliminates artifacts, enhancing image sharpness and clarity, and leading to a more accurate and realistic reconstruction, particularly in high-frequency regions.

\begin{table}[t]
    \centering
    \footnotesize 
    \renewcommand{\arraystretch}{1.3} 
    \begin{tabular}{>{\centering\arraybackslash}p{0.2\columnwidth} | >{\centering\arraybackslash}p{0.13\columnwidth} >{\centering\arraybackslash}p{0.13\columnwidth} | >{\centering\arraybackslash}p{0.13\columnwidth} >{\centering\arraybackslash}p{0.13\columnwidth}}
        \toprule
        & \multicolumn{2}{c|}{PSNR$\uparrow$} & \multicolumn{2}{c}{SSIM$\uparrow$} \\
        Filter & w/ & w/o & w/ & w/o \\
        \hline
        Factory & 37.75 & 33.08 & .9646 & .9230 \\
        Vender & 36.00 & 31.60 & .9641 & .9334 \\
        Flamingo & 31.33 & 26.38 & .9022 & .7836 \\
        Roundabout & 31.07 & 22.38 & .9222 & .7331 \\
        \bottomrule
    \end{tabular}
    \caption{\textbf{The results of ablation experiment validate the high-frequency filter.} The results show that the introduction of high-frequency filters significantly improves the image quality.}
    \label{tab:ablation}
\end{table}

%% file: sec/5_conclusion.tex
\section{Conclusion}

This paper proposes a novel method for recovering dynamic 3D scene representations from a single snapshot conpressive image, which is the first to introduce an dynamic explicit representation in this task, extending its application to dynamic scenes. We propose a transformation network to substitute for directly optimizing the camera poses and a high-frequency filter to eliminate the artifacts generated during the transformation. Different from previous works, our method can adequately reconstructing dynamic scenes from compressed image, and provides a new idea for optimizing camera poses with the absence of camera poses and pre-trained 3DGS representations. To access the effectiveness of SCIGS, extensive comparative experiment are conducted against the existing state-of-the-art SCI image recovery methods and SCI image-based reconstruction methods both in static scenes and dynamic scenes. And the result demonstrate the superior performance of our method, especially in dynamic scenes.

Due to the advantages of SCI technology in terms of storage cost, and the superiority and scalability of our 3DGS-based framework demonstrated in dynamic scenes, our method has great potential for low-cost and fast incremental reconstruction in high-speed dynamic scenes, such as autonomous driving.

%% file: sec/X_suppl.tex
\clearpage
\setcounter{page}{1}
\maketitlesupplementary
\renewcommand{\thesection}{\Alph{section}}
\setcounter{section}{0}
\renewcommand{\thefigure}{\Alph{figure}} 
\renewcommand{\thetable}{\Alph{table}}   
\setcounter{figure}{0} 
\setcounter{table}{0}  

In this supplementary material, additional experiments are conducted on datasets from dynamic scenes and static scenes, comparing our SCIGS against current state-of-the-art SCI decoding methods (GAP-TV\cite{yuan2016generalized}, PnP-FFDNet\cite{yuan2020plug}, PnP-FastDVDNet\cite{yuan2021plug} and EfficientSCI\cite{wang2023efficientsci}) and state-of-the-art SCI image-based reconstruction method(SCINeRF\cite{li2024scinerf}).

\section{Additional Experiments}
\label{sec:a_experiments}

\subsection{Experiment Setup}
To further validate the effectiveness of our method in dynamic scenes, additional qualitative and quantitative experiments were conducted on five datasets from dynamic scene ($Bear$, $Roundabout$, $Flamingo$, $Turn$ and $Dance$). For fair comparisons, we fine-tuned EfficientSCI~\cite{wang2023efficientsci} with the masks used in our datasets. Additionally, the results of qualitative experiments conducted under all static scene datasets ($Factory$, $Tanabata$, $Vender$, $Cozy2room$, $hotdog$ and $airplants$) are also presented in this supplementary material. For a better quantitative comparison, we also present the results of the experiments with static scenes, which are shown in Table~\ref{tab:a_static}.

\subsection{Result and Analysis}
\begin{figure*}[t]
\centering
\includegraphics[width=1.0\textwidth]{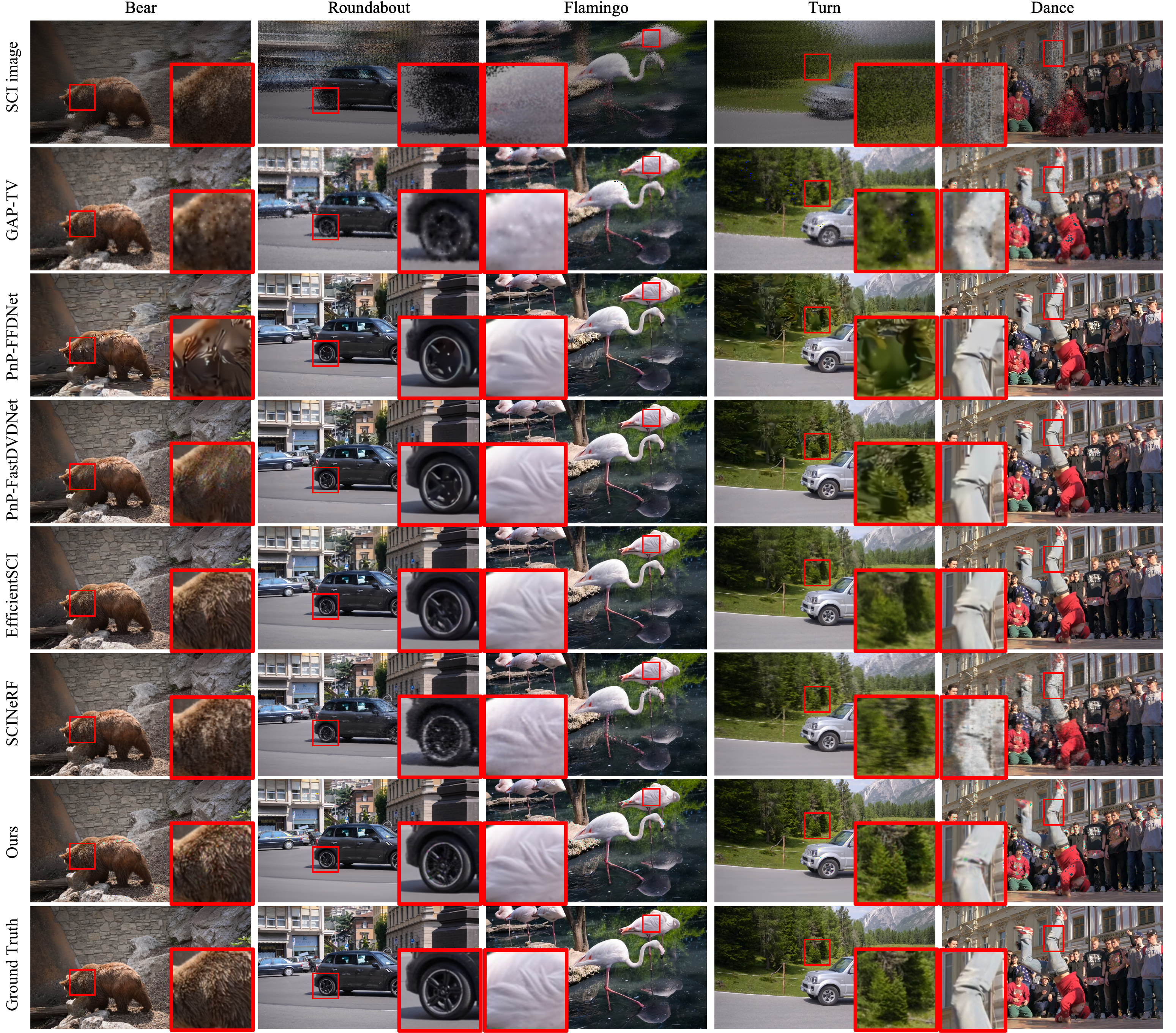}
\caption{\textbf{Qualitative evaluations on the datasets from dynamic scenes.}  From left to right shows the results for five dynamic scenes including $Bear$, $Roundabout$, $Flamingo$, $Turn$ and $Dance$. The experiments show that our method achieves superior performance in dynamic scenes.}
\label{fig:s_dynamic}
\end{figure*}

The results of the qualitative and quantitative experiments in dynamic scenes are shown in Fig.~\ref{fig:s_dynamic} and Table~\ref{tab:a_dynamic}, respectively. These results provide empirical evidence for the effectiveness of our SCIGS in reconstructing dynamic scenes from single compressed images. It is also worth noting that the metrics of our method do not exceed EfficientSCI in $Dance$. The observation can be attributed to the fact that our method recovers images by reconstructing the underlying scene. The images in the $Dance$ dataset have dynamic blur, which leads to the loss of structural information, so SCIGS cannot accurately reconstruct this part of the scene, leading to a degradation in image quality. In contrast, as a traditional SCI image decoding methods, EfficientSCI uses only 2D image information without considering the structural consistency, and thus outperforms our method in this scene.

\begin{figure*}[t]
\centering
\includegraphics[width=0.97\textwidth]{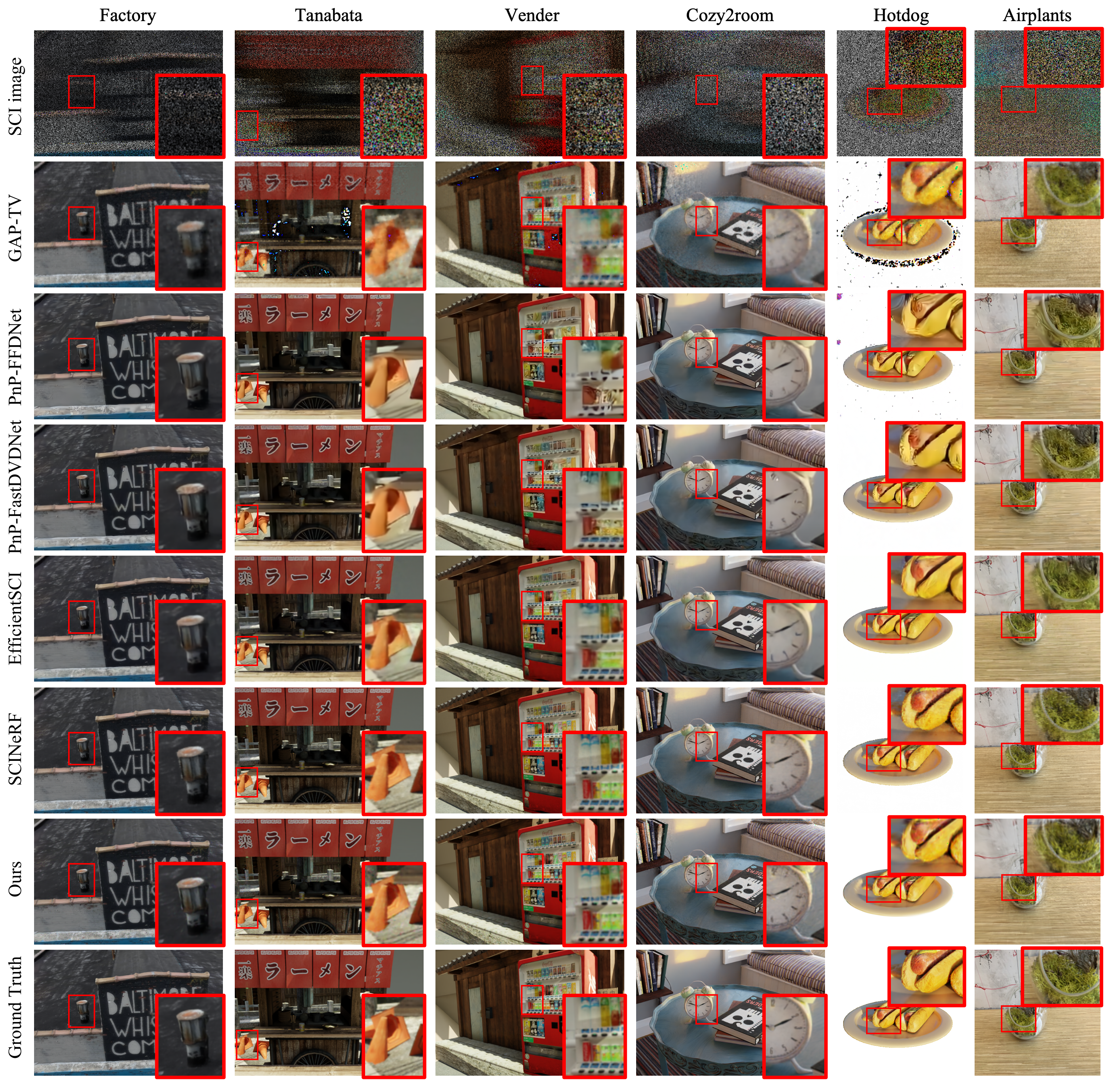}
\caption{\textbf{Qualitative evaluations on the datasets from static scenes.}  From left to right shows the results for five static scenes including $Factory$, $Tanabata$, $Vender$, $Cozy2room$, $Hotdog$ and $Airplants$. The experiments show that our method achieves comparable image recovery performance from a single compressed image in static scenes.}
\label{fig:s_static}
\end{figure*}

As shown in Fig.~\ref{fig:s_static} and Table~\ref{tab:a_static}, the proposed SCIGS shows comparable image recovery performance on static scene. In addition, we note that SCIGS outperforms existing methods in the reconstruction of the parts with rich textures and characters, which cannot be directly observed from metrics.

\begin{table*}[t]
\centering
\renewcommand{\arraystretch}{1} 
\resizebox{\textwidth}{!}{
    \begin{tabular}{c|ccc|ccc|ccc|ccc|ccc}
        \toprule
            & \multicolumn{3}{c|}{Bear} & \multicolumn{3}{c|}{Roundabout} & \multicolumn{3}{c|}{Turn} & \multicolumn{3}{c|}{Flamingo} & \multicolumn{3}{c}{Dance} \\
            & PSNR$\uparrow$ & SSIM$\uparrow$ & LPIPS$\downarrow$ & PSNR$\uparrow$ & SSIM$\uparrow$ & LPIPS$\downarrow$ & PSNR$\uparrow$ & SSIM$\uparrow$ & LPIPS$\downarrow$ & PSNR$\uparrow$ & SSIM$\uparrow$ & LPIPS$\downarrow$ & PSNR$\uparrow$ & SSIM$\uparrow$ & LPIPS$\downarrow$ \\
        \hline
        GAP-TV\cite{yuan2016generalized} & 22.63 & .5698 & .3734 & 22.26 & .6976 & .3823 & 25.28 & .6774 & .3437 & 23.68 & .6986 & .3404 & 22.20 & .6981 & .3953 \\
        PnP-FFDNet\cite{yuan2020plug} & 21.91 & .6569 & .3822 & 25.80 & .8727 & .1314 & 26.93 & .8598 & .2661 & 25.50 & .8206 & .2000 & 22.29 & .8284 & .1987 \\
        PnP-FastDVDNet\cite{yuan2021plug} & 26.77 & .8561 & .1413 & 27.01 & .8938 & .1006 & 27.58 & .8723 & .2090 & 29.27 & .8978 & .0994 & \underline{28.10} & \underline{.9465} & \underline{.0569}\\
        EfficientSCI\cite{wang2023efficientsci} & \underline{29.26} & \underline{.9099} & \underline{.0710} & \underline{28.45} & \underline{.9110} & \underline{.0876} & \underline{29.03} & \underline{.8934} & \underline{.1617} & \underline{31.03} & \textbf{.9247} & \underline{.0668} & \textbf{31.55} & \textbf{.9677} & \textbf{.0412} \\
        SCINerf\cite{li2024scinerf} & 26.57 & .7974 & .1192 & 26.02 & .8394 & .1265 & 25.68 & .6596 & .2330 & 26.78 & .7954 & .1207 & 22.78 & .6960 & .2737 \\
        \hline
        SCIGS(ours) & \textbf{30.44} & \textbf{.9137} & \textbf{.0548} & \textbf{31.07} & \textbf{.9222} & \textbf{.0729} & \textbf{31.78} & \textbf{.8951} & \textbf{.0953} & \textbf{31.33} & \underline{.9022} & \textbf{.0533} & 27.89 & .9096 & .0580 \\
        \bottomrule
    \end{tabular}
}
\caption{\textbf{Quantitative SCI image reconstruction comparisons on the dynamic datasets.} The results demonstrate that our method surpasses the current SCI decoding methods and 3D reconstruction methods for SCI image on datasets from dynamic scenes. The best results are shown in bold and the second-best results are underlined.}
\label{tab:a_dynamic}
\end{table*}

\begin{table*}[t]
\begin{center}
\huge
\renewcommand{\arraystretch}{1.0} 
\resizebox{\textwidth}{!}
{
    \begin{tabular}{c|ccc|ccc|ccc|ccc|ccc|ccc}
        \toprule
            & \multicolumn{3}{c|}{Cozy2room} & \multicolumn{3}{c|}{Tanabata} & \multicolumn{3}{c|}{Factory} & \multicolumn{3}{c|}{Vender} & \multicolumn{3}{c|}{Airplants} & \multicolumn{3}{c}{Hotdog} \\
            & PSNR$\uparrow$ & SSIM$\uparrow$ & LPIPS$\downarrow$ & PSNR$\uparrow$ & SSIM$\uparrow$ & LPIPS$\downarrow$ & PSNR$\uparrow$ & SSIM$\uparrow$ & LPIPS$\downarrow$ & PSNR$\uparrow$ & SSIM$\uparrow$ & LPIPS$\downarrow$ & PSNR$\uparrow$ & SSIM$\uparrow$ & LPIPS$\downarrow$ & PSNR$\uparrow$ & SSIM$\uparrow$ & LPIPS$\downarrow$ \\
        \hline
        GAP-TV\cite{yuan2016generalized} & 21.77 & .4321 & .6031 & 20.42 & .4264 & .6250 & 24.05 & .5666 & .5149 & 20.00 & .3678 & .6882 & 22.85 & .4057 & .4986 & 22.35 & .7663 & .3179 \\
        PnP-FFDNet\cite{yuan2020plug} & 28.98 & .8916 & .0984 & 29.17 & .9032 & .1197 & 31.75 & .8977 & .1142 & 28.70 & .9235 & .1315 & 27.79 & .9117 & .1817 & 29.00 & \underline{.9765} & .0511 \\
        PnP-FastDVDNet\cite{yuan2021plug} & 30.19 & .9132 & .0793 & 29.73 & .9333 & .0980 & 32.53 & .9165 & .1055 & 29.68 & .9395 & .1043 & 28.18 & .9092 & .1757 & 29.93 & .9728 & .0522 \\
        EfficientSCI\cite{wang2023efficientsci} & 31.47 & \underline{.9327} & .0476 & 32.30 & \underline{.9587} & .0600 & 32.87 & .9259 & .0709 & 33.17 & .9401 & .0456 & \underline{30.13} & \textbf{.9425} & \underline{.1129} & \underline{30.75} & .9568 & \underline{.0461} \\
        SCINerf\cite{li2024scinerf} & \underline{33.23} & \textbf{.9492} & \underline{.0445} & \underline{33.61} & \textbf{.9638} & \underline{.0374} & \underline{36.60} & \underline{.9638} & \textbf{.0221} & \textbf{36.40} & \textbf{.9840} & \underline{.0298} & \textbf{30.69} & \underline{.9335} & \textbf{.0728} & \textbf{31.35} & \textbf{.9878} & \textbf{.0310} \\
        \hline
        SCIGS(ours) & \textbf{33.78} & .9191 & \textbf{.0423} & \textbf{35.12} & .9580 & \textbf{.0271} & \textbf{37.75} & \textbf{.9646} & \underline{.0291} & \underline{36.00} & \underline{.9641} & \textbf{.0192} & 27.18 & .7267 & .3003 & 29.31 & .9369 & .0809 \\
        \bottomrule
    \end{tabular}
}
\caption{\textbf{Quantitative SCI image reconstruction comparisons on the static datasets.} The results demonstrate that our method outperforms or approaches the existing image reconstruction methods and 3D reconstruction methods for SCI image on most datasets from static scenes. The best results are shown in bold and the second-best results are underlined.}
\label{tab:a_static}
\end{center}
\vspace{-20pt}
\end{table*}

